\documentclass{article} 
\pdfoutput=1
\usepackage{nips15submit_e,times}
\usepackage{hyperref}
\usepackage{url}
\usepackage{wrapfig,graphicx,subfig}
\graphicspath{./}

\title{Detecting Driveable Area for Autonomous Vehicles}

\author{
Niral Shah \\
Department of ECE\\
Duke University\\
\texttt{ns247@duke.edu} \\
\And
Ashwin Shankar\\
Department of ECE \\
Duke University \\
\texttt{as820@duke.edu} \\
\And
Jae-hong Park \\
Department of BME \\
Duke University \\
\texttt{jp361@duke.edu} \\
}

\begin{document}

\maketitle
\begin{abstract}
Autonomous driving is a challenging problem where there is currently an intense focus on research and development. Human drivers are forced to make thousands of complex decisions in a short amount of time,quickly processing their surroundings and moving factors. One of these aspects, recognizing regions on the road that are driveable is vital to the success of any autonomous system. This problem can be addressed with deep learning framed as a region proposal problem. Utilizing a Mask R-CNN trained on the Berkeley Deep Drive (BDD100k) dataset, we aim to see if recognizing driveable areas, while also differentiating between the car's direct (current) lane and alternative lanes is feasible. 
\end{abstract}

\section{Problem Description }
To enable autonomous driving, researchers have broken the problem down into various sub-problems, which include road object detection,  segmentation or breaking a 2D image into depth based layers, lane marker detection, and identifying driveable areas directly in front of the vehicle. The latter, identifying driveable areas within an image, is the focus of this paper. The goal is to use annotated training data from the Berkeley Deep Drive (BDD) dataset and test whether a deep learning network can successfully identify driveable regions in a scene. 
\newline The importance of recognizing driveable areas is arguably paramount in the process of developing an autonomous driving solution. As humans, recognizing driveable areas is typically a trivial problem, we can clearly process and visualize the road directly in front of us with our eyes and brain. We know exactly what’s a road, what’s a sidewalk, where there are specialized lanes for bicycles or buses. Therefore if vehicles are to supersede a human’s ability to drive, they need to be able to solve this basic problem. Nevertheless, the problem remains a challenge to researchers.
\newline As an area of intense research, there does exist some literature on detecting driveable regions. However  many researchers have utilized high tech equipment like 3D Laser Imaging, Detection And Ranging (LIDAR) scanners as well as traditional radars to create a more holistic picture of the vehicle’s surroundings to improve their decision making. Alternatively several technology companies including Tesla have approached the problem by assuming structured roads and simply looking at line markings for informing autonomous decision making. As a result, literature related to proposing regions within an video sequence does not appear to have been publicly disclosed. Though researchers at Facebook AI, have developed a region proposing convolutional neural network, called Mask R-CNN[1] that will be applicable to this problem. Mask R-CNN develops upon a popular convolutional network known as Faster R-CNN [2] by taking bounding boxes and applying segmentation within the bounding box to get a specific shape. This should allow us to classify driveable regions within a frame and evaluate the results on the training data.

\section{Data Description}
\begin{figure}
{%
\setlength{\fboxsep}{0pt}%
\setlength{\fboxrule}{1pt}%
\fbox{\includegraphics[width=0.7\linewidth]{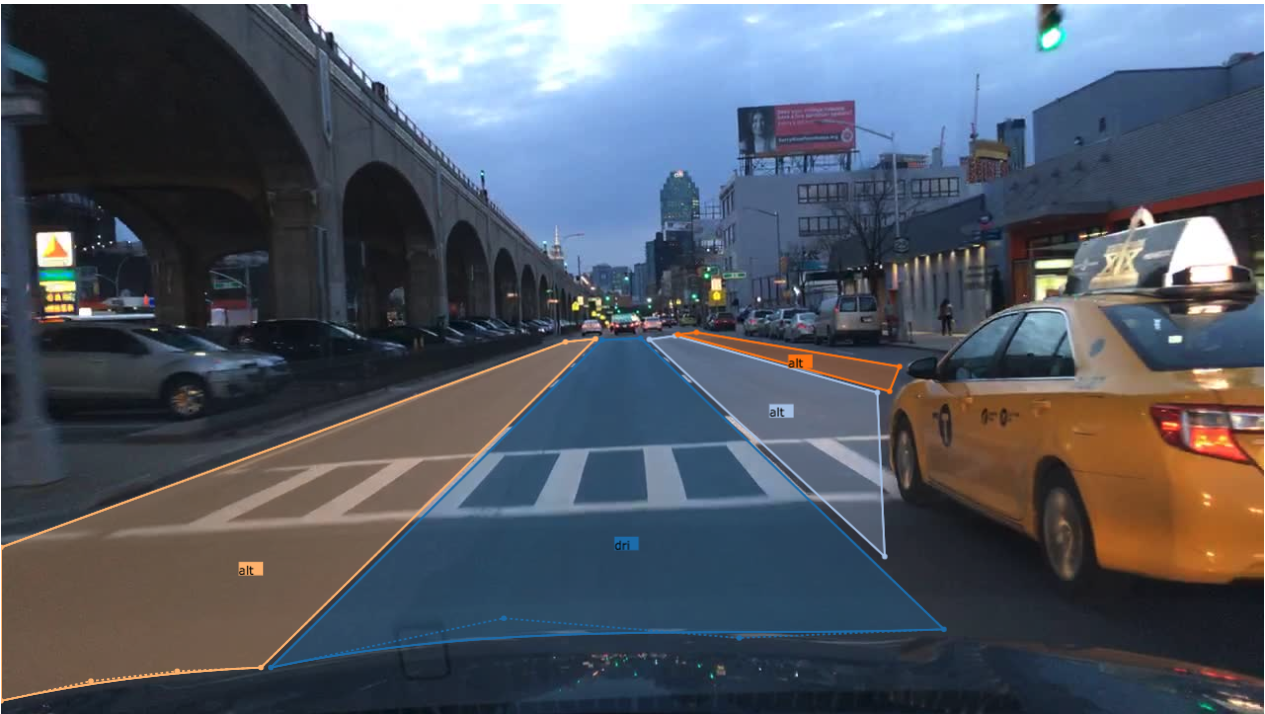}}
}%
\centering
\caption{Example of an Image from Dataset Annotated with Driveable Areas[3]}
\end{figure}
In the project, BDD100k, a database of diverse driving video (http://bdd-data.berkeley.edu/) is being used. The database consists of 100k road view images recorded from cameras implemented in the cars [3]. The images here provide a front view while the vehicles are driving, for example see figure 1. The images were acquired in four regions, New York, San Francisco, Berkeley, and the Bay Area, where traffic conditions are so complex.  Images are allocated for training, validation, and test in quantities of 70k, 10k, 20k respectively. For the training and validation groups JSON files containing annotations and labels are provided. They contain various weather and different time conditions (sunny/ rainy/ snowy, daytime/ nighttime/ dusk or dawn). The test set does not have any publicly available labels. Since the dataset covers a wide variety of conditions that are likely occur while driving, our final result should be well generalized.

\section{Preprocessing}
As described in the previous section, the BDD100k dataset provides three sets of data: training,validation and test of which there are 70k, 10k, and 20k respectively. However, since the test set does not have any provided annotations or labels it cannot be used for the purposes of training or evaluation. Thus this project focuses on the remaining 80k images from the training and the validation sets. The annotations file provided contains the coordinates of the polygons enclosing the driveable areas. Each image could have multiple polygons or could potentially have none. Thus preprocessing was done by generating a new annotations file in JSON files that contained a list of all the x and y coordinates for each polygon and which contained only images in which driveable areas are present. This resulted in approximately 4.5\% of images being dropped in the training and validation sets respectively. 
Subsequently, masks were obtained and generated as bitmaps from the annotations stored in JSON files. The preprocessing was careful to preserve the difference between regions that were considered "direct" or the vehicle's current lane and "alternative" or adjacent lanes. Generating these bitmap masks is necessary as it is a part of the expected input for the Mask R-CNN.

\section{Core Methods}
In this problem the data provided are high resolution dash cam images taken from a moving vehicle across the country and in various settings. As a result the process of recognizing driveable areas using this dataset turns into an image classification problem with region proposals. 

Convolution Neural Networks (CNN) have demonstrated a great capability of classification on images. The performances of several models with CNN have similar or even out performed human recognition [4,5,6]. However, we cannot directly apply these models to our goal of detecting driveable areas in  autonomous driving application, given that each image has several object with various locations, which is beyond canonical classification problems. Therefore, another strategy should be employed to address problem.

To localize and detect objects in an image, Region-based Convolutional Neural Network (R-CNN) was developed (Figure 2)[7]. R-CNN uses Alexnet [8] for transfer learning and duplicates the model into two distinct models of classification (reduced to 21 classes from 1000) and regression for object detection and the coordinates of objects (x,y,w,h for a box) respectively. R-CNN uses Selective Search algorithm [9] to extract region proposals up to 2000 to feed the models. Each region proposal should be resized to the same dimensiomns before feeding it to the CNN. At the last step, features from the CNN are classified by support vector machine (SVM) into the predefined categories. The R-CNN demonstrates improved performance on object detection. 
However, R-CNN has several problems. Since R-CNN applies CNN 2000 times for each region proposal (2k), the runtime of the model is really slow (Training in 84 hours and Testing one image in 49 seconds). In addition, the training of the model is extremely difficult, due to the multi-level pipelines, including the CNN, the classifiers, and the regressor.
\begin{wrapfigure}{l}{0.7\textwidth}
    \centering
    \includegraphics[width=0.7\textwidth]{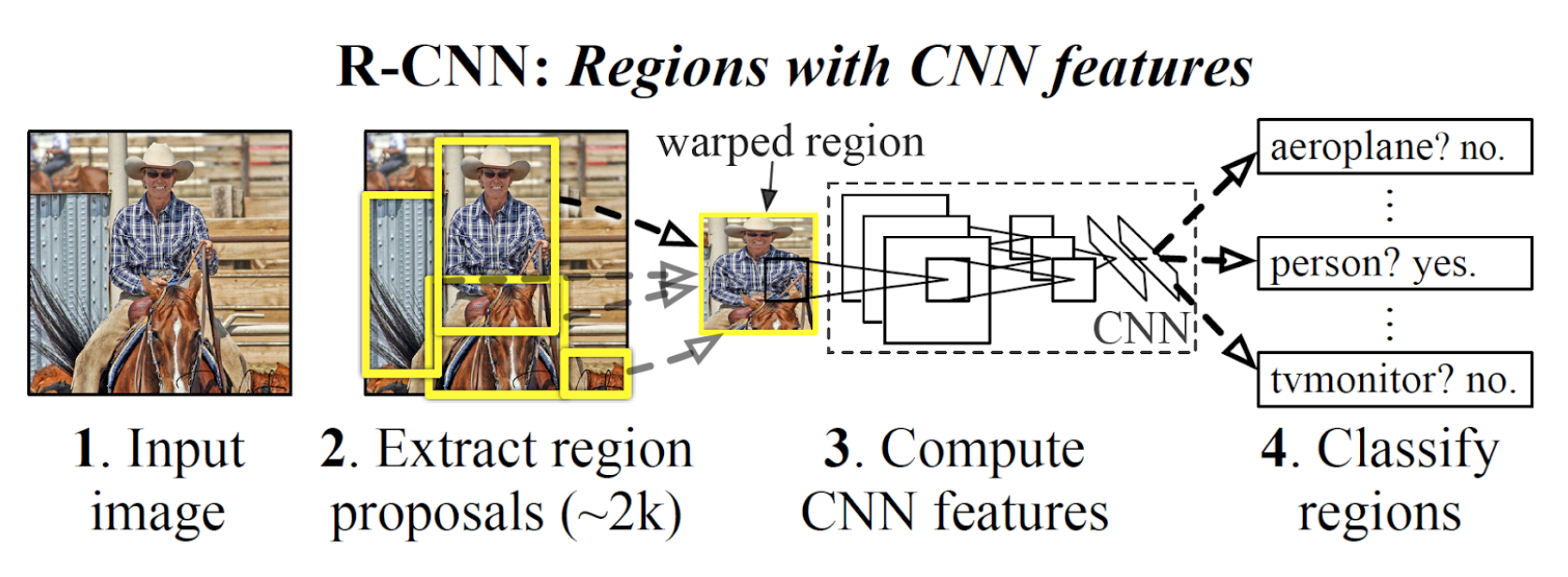}
    \caption{[7]}
\end{wrapfigure}

Fast R-CNN facilitates its speed by combining three pipelines into one[10] (with respect to the improvement over R-CNN). Instead of applying the CNN for every region proposal, Fast R-CNN takes an image to the one CNN to generates features. The features then are equally cropped in size by using ROI pooling and the region proposals (2000 from selective search in the beginning). The fully connected layers with the ROI features  are fed into two output layers for classification and regression. Therefore, Fast R-CNN jointly trains for classification and regression for bounding boxes.
The runtime for training and testing of Fast R-CNN are 8 hours and 2.3 seconds respectively.(84 hours and 49 seconds with R-CNN). Ren et al, have argued that the most time consuming step in fast R-CNN is the region proposal section. 
To address the bottleneck, they propose Faster R-CNN by integrating region proposals into the deep neural network as Region Proposal Network (RPN) to minimize the computation cost.
RPN takes the output feature maps from the CNN and returns its output with rectangular objects and objectness scores for region proposals. 
To this end, they predict region proposals while sliding a window by applying anchor boxes that are references for common aspect ratios (e.g. 1:1, 2:1,1:2). 

The runtime of Faster R-CNN is 0.2 seconds for testing, which seems to be reasonable for real-time use for autonomous driving application. The last thing we need to consider on our goal is shapes of detected objects. Images of road from car cameras are not rectangles, but more of trapezoids, sometimes not even defined. Therefore, in order to detect driveable areas in images a model requires instance segmentation.
Mask R-CNN is an alternative approach that can generate instance segmentation of objects, which is suitable for detecting a driveable region that is not a rectangle. Mask R-CNN combines Faster R-CNN with Fully Convolutional Networks (FCN) to predict pixel segmentation on each classified object. FCN [11] has been developed for semantic segmentation. 
FCN replaces fully connected layer with a fully convolutional network by applying 1x1 convolution filter, thus  maintaining spatial information of features while reducing computational costs. In addition, Mask R-CNN improves the accuracy of feature map location. Previously, ROI pooling is used to get a feature map from ROI, which inevitably creates a little misalignment between the features and the ROIs. In masking segmentation, it is important to minimize this misalignment. To fix the misalignments, Mask R-CNN proposes ROIAlign that reliably maintains the original location by using 2d linear interpolation while calculating feature maps from ROI.

The problem addressed here is object instance segmentation of drivable areas. At this moment, Mask R-CNN is the best method for this problem. There are many hyperparameters here such as anchor scales, anchor ratios, and the parameters  for stochastic gradient descent (SGD) with momentum. The anchor  ratios are the same as that in the Faster R-CNN paper. The default anchor scales used are a fit for our problem. The SGD parameters used are the default parameters with a learning rate of 0.002 and a momentum of 0.9. The metric used for measuring performance is mean average precision (mAP). This is obtained by getting the area under a modified precision-recall graph where each recall value is mapped to the max precision obtained for recall values from itself to one.This metric is important as it accounts for the objects proposed as well as the ranking of the same objects based on descending order of predicted probabilities.

\section{Final Results}
\begin{figure}
\centering
\includegraphics[width=1\linewidth]{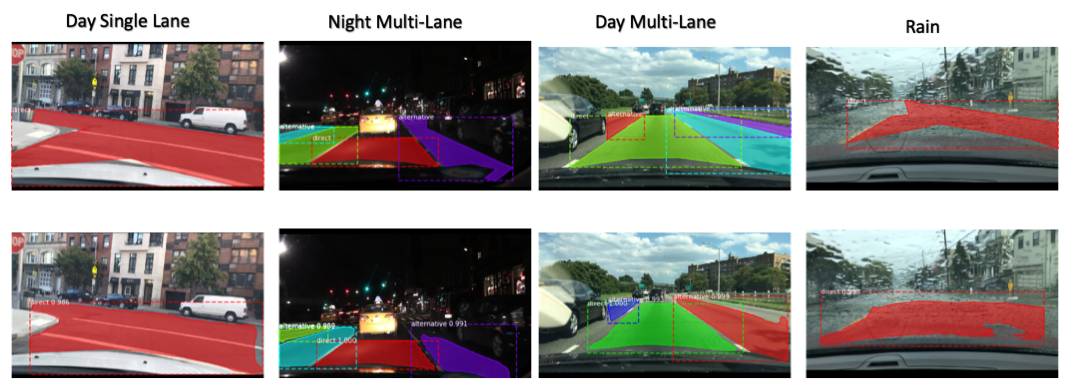}
\caption{Top Row: Ground-Truth Annotations, Bottom Row: Model Detections}
\end{figure}
A Mask R-CNN model was successfully trained to identify driveable areas including both direct and alternative lanes within a single scene. As can be seen in Figure 3, the model appears to be a robust solution to this critical component of an autonomous vehicles.

For the initial development, training was done in two steps. Initially all layers were frozen except the final head layers. Weights were initialized from the training done previously on the Microsoft COCO dataset [12]. The COCO dataset has a wide variety of objects and corresponding masks and serves as a good general base for the weights. 

Subsequently, for preliminary results, training was done on the reduced validation set of 9,546 images due to time and computational restrictions. A single GPU running on a virtual machine (VM) on Google Cloud was used. The model was trained for 666 epochs with a batch size of 100 images on the final head layers. Observing the results, the head training saturated with a loss of about 0.53. This was a great starting point, as by simply training a handful of layers, the network was able to significantly improve it's performance and was ready for the next phase, fine-tuning. 

\begin{figure}
\centering
\includegraphics[width=1\linewidth]{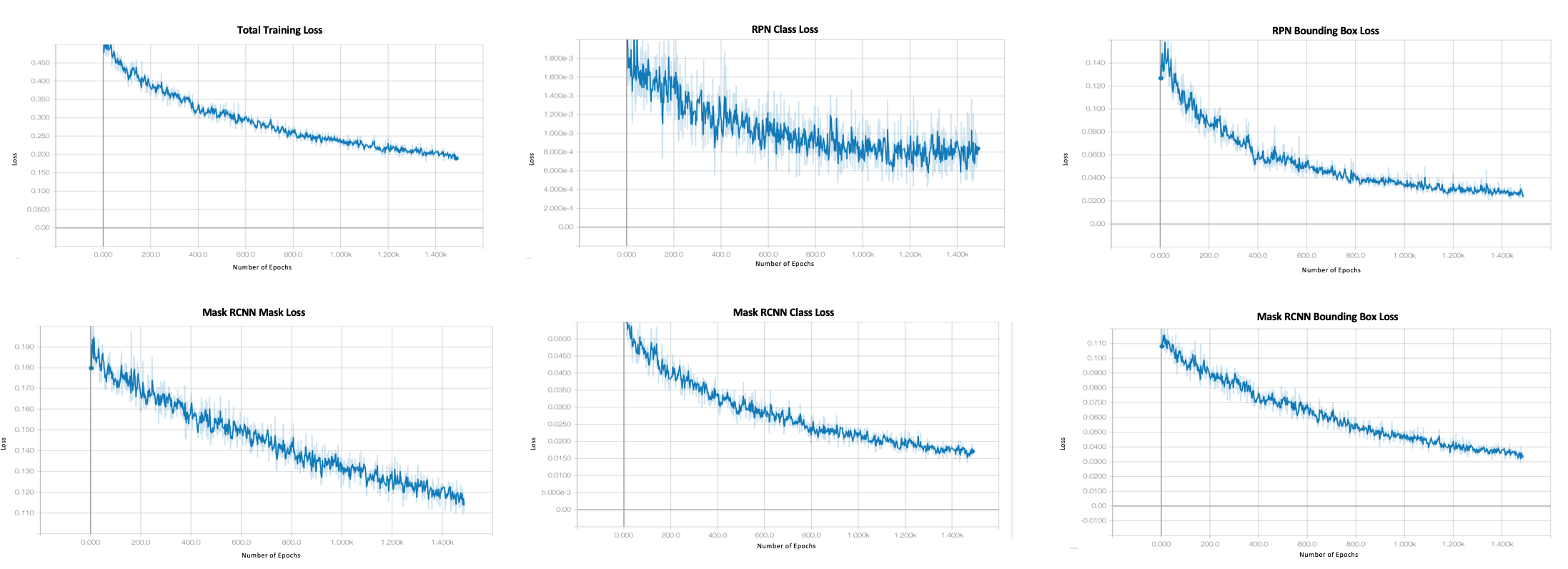}
\caption{Training Loss Graphs, Top Left Total Loss, the other Five Losses sum up to Total Loss}
\end{figure}

In the fine-tuning phase, the original plan was to train the network for 2000 epochs with a batch size of 100 images each. However, observing the loss, as can be seen in Figure 4, at approximately 1400 epochs, the loss began to saturate and stabilize around 0.2. Due to the large size of the images being trained on, and the limited computational resources available, this process could not be repeated with other techniques to see if loss could be reduced further. The final result achieved a Mean Average Precision of 0.7938 on the validation set and 0.9263 on the training set. The metric of mean average precision, is a bit abstruse as it may not fully reveal the pitfalls of the network. For example the network may suggest the correct classes but the proposed regions may be smaller or larger than the annotated regions. In such a case, one can increase the threshold  of Intersection over Union(IoU) which might then reduce mAP. Furthermore, the annotated regions are all flat polygonal shapes, while Mask R-CNN as a network built for instance segmentation does not typically generate polygonal masks. 

\begin{figure}[htp]
\centering
\includegraphics[width=1\linewidth]{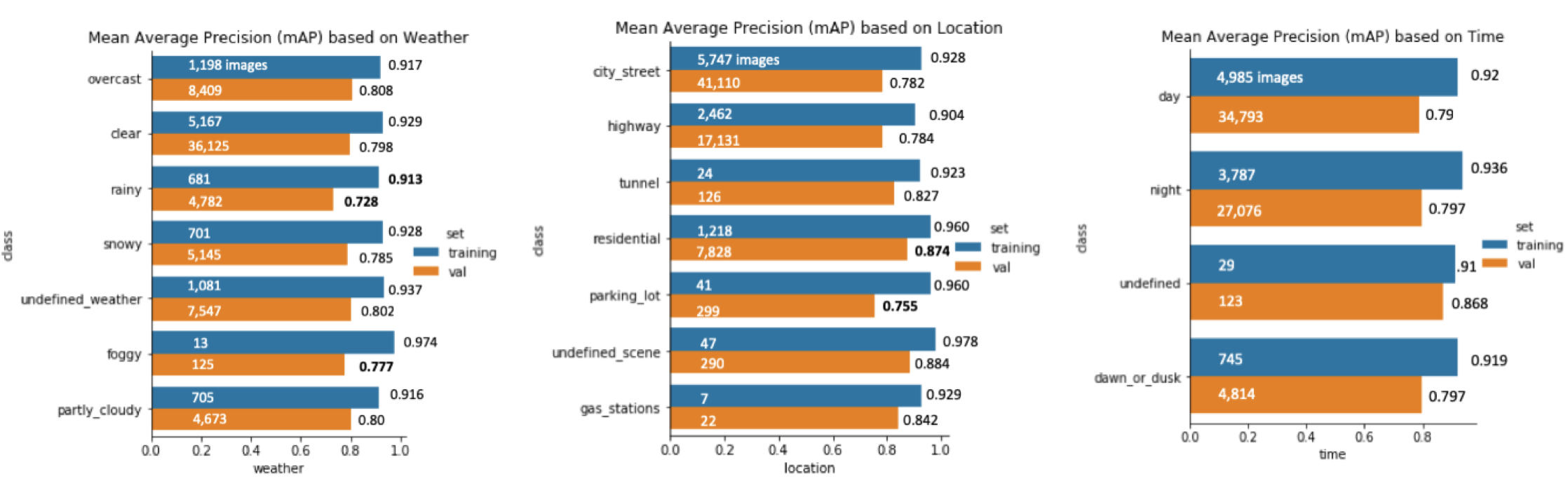}
\caption{Model Performance Based on Weather, Location, and Time Conditions}
\label{fig:figure3}

\end{figure}

The BDD Dataset provides for most images, information about the conditions upon which an image was taken. The conditions can be broken down into three categories, Weather, Location, and Time. Figure 5 highlights the mean average precision (mAP) under each of these conditions for both the Validation and Training Sets. When breaking the results down into these categories interesting results emerge. For example, its clear that the model struggles with images in rainy conditions and performs very well in residential locations. Many other categories, like foggy conditions, represent less than five percent of the training set, and thus should not be considered in the analysis. Another interesting result, was that the model performed equally well in all time conditions especially when comparing day and night conditions. This ran counter to the expectation that night results would likely do worse due difficulties in viewing the lane markers. However it appears the model comfortably was able to learn. Finally the most interesting result was that despite the relatively small training sizes for each of these categories, the model was still able to perform well.

\section{Conclusion and Future Work}

\begin{figure}
\centering
\includegraphics[width=0.6\linewidth]{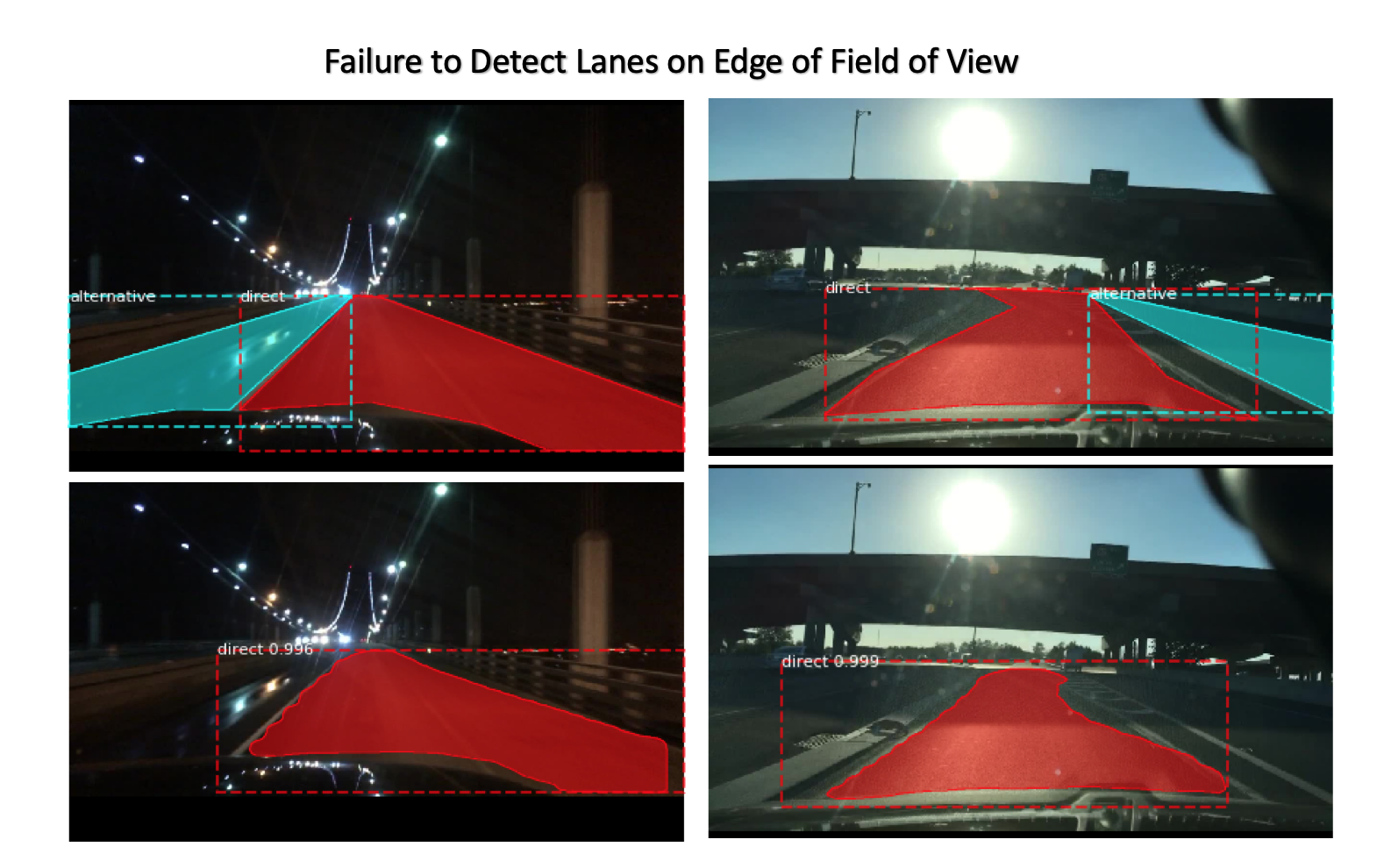}
\caption{An example of one of the network's pitfalls}
\end{figure}

After training a Mask R-CNN on model on the BDD Dataset, the model performs relatively well achieving a loss of 0.2 and a mean average precision (mAP) of 0.79. In this process, several unique decisions were made. The biggest choice was to use a relatively small sample size of 9,546 images and validate on a larger set. The hope was to gain early insight into how well Mask R-CNN could perform on the application of driveable areas. The next unique choice was to initialize the network with COCO weights and train only the head layers to reduce loss as much as possible, stopping once the loss saturated at around 0.5. Subsequently, this loss was further reduced by fine-tuning all the layers, achieving the final result. Since this dataset is relatively new, there is no previous work or literature to compare the results to.  For the development and training we referred to the github repository setup by the Mask R-CNN creators (Matterport: \url{https://github.com/matterport/Mask_RCNN}).

Analyzing the results revealed that the model performs well under most driving conditions. Interestingly, the model performs significantly better in residential locations, perhaps due to the similarities in images across the class. Some cases which the network appears to struggle on include rainy weather having a slightly higher disparity between validation and training results compared to other weather conditions. This could potentially be addressed by utilizing a different type of neural network better suited for handling occluded scenes. Some cursory research reveals that there are some proposed methods for rain removal from images using convolutional neural networks[13]. 

By most metrics described here it appears the network is doing very well and can handle most scenarios. Some scenarios like rain may need to be handled by other types of networks as a form of preprocessing and subsequently fed to a Mask R-CNN model. In addition, as seen in Figure 6, simply through observation, it appears that in many multi-road cases the network struggles to find alternative lanes on the far right and far left corners of the image. Since the dataset does not distinguish these images, analysis could not be completed on how many images like these are affecting the mean average precision. This would also be an area the team will be continuing to explore. Another critical pitfall of this current model was that it does not handle scenarios in which the image does not contain any driveable regions. While the initial training set included some images of this category,  a decision was made to ignore these images. However a real-world application of this model would require this case to be handled as well. Finally, as mentioned, the network was trained on a relatively small training set of 9,546 images compared to a validation set 66,806 images. Due to the nearly 70k images available it could be more beneficial to train on an even larger split to see if a better result set could be obtained. These are just a few of the possible future steps that could be taken. 

Overall, the team is very excited by the results and are impressed that in a relatively short time with this dataset, a critical component of an autonomous vehicle system was built!


\begin{thebibliography}{9}
\bibitem{mask-r-cnn}
K. He, G. Gkioxari, P. Dollar, and R. Girshick. Mask R-CNN.arXiv:1703.06870 (https://arxiv.org/abs/1703.06870)

\bibitem{fastercnn}
S. Ren, K. He, R. Girshick, and J. Sun. Faster R-CNN: Towards real-time object detection with region proposal networks. In TPAMI , 2017

\bibitem{bdd100k}
BDD100K: A Diverse Driving Video Database with Scalable Annotation Tooling (https://arxiv.org/pdf/1805.04687.pdf)


\bibitem{deepConv}
Szegedy, C., Liu, W., Jia, Y., Sermanet, P., Reed, S., Anguelov, D., ... and Rabinovich, A., "Going deeper with convolutions," Proceedings of the IEEE conference on computer vision and pattern recognition, 1-9 (2015).

\bibitem{inceptionV4}
Szegedy, C., Ioffe, S., Vanhoucke, V. and Alemi, A. A., "Inception-v4, Inception-ResNet and the Impact of Residual Connections on Learning," AAAI, 4278-4284 (2017).

\bibitem{deepConv}
He, K., Zhang, X., Ren, S. and Sun, J., "Deep residual learning for image recognition," In Proceedings of the IEEE conference on computer vision and pattern recognition, 770-778 (2016).

\bibitem{r-cnn}
Girshick, R., Donahue, J., Darrell, T. and Malik, J., "Region-based convolutional networks for accurate object detection and segmentation, " IEEE transactions on pattern analysis and machine intelligence, 142-158 (2016).

\bibitem{imagenet}
Krizhevsky, A., Sutskever, I. and Hinton, G. E., "Imagenet classification with deep convolutional neural networks," In Advances in neural information processing systems, 1097-1105 (2012).

\bibitem{selectiveSearch}
Uijlings, J. R., Van De Sande, K. E., Gevers, T. and Smeulders, A. W., "Selective search for object recognition, " International journal of computer vision, 104(2), 154-171(2013).

\bibitem{fastRCNN}
Girshick, R., "Fast r-cnn, " In Proceedings of the IEEE international conference on computer vision, 1440-1448 (2015).

\bibitem{semanticSeg}
Long, J., Shelhamer, E. and Darrell, T., "Fully convolutional networks for semantic segmentation," In Proceedings of the IEEE Conference on Computer Vision and Pattern Recognition, 3431-3440 (2015).

\bibitem{coco}
Tsung-Yi Lin, Michael Maire, Serge Belongie, Lubomir Bourdev, Ross Girshick, James Hays, Pietro Perona, Deva Ramanan, C. Lawrence Zitnick, Piotr Dollár, "
Microsoft COCO: Common Objects in Context" arXiv e-prints, arXiv:1405.0312 [cs.CV], 2014

\bibitem{rain-removal}
Chen Jie, Tan Cheen-Hau, Hou Junhui, Chau Lap-pui, Li He, 
"Robust Video Content Alignment and Compensation
for Rain Removal in a CNN Framework", arXiv e-prints
arXiv: 803.1043 [cs.CV], 2014
\end{thebibliography}
\end{document}